\documentclass[runningheads]{llncs}
\usepackage{graphicx}
\usepackage{physics}
\usepackage{amsfonts}
\begin{document}
\title{Prostate Tissue Grading with Deep Quantum Measurement Ordinal Regression}
%
\titlerunning{Prostate Tissue Grading with DQMOR}
%

\author{ Santiago Toledo-Cortés
\and Diego H. Useche
\and Fabio A. González
}
\authorrunning{S. Toledo-Cortés et al.}

%
\institute{MindLab Research Group, Universidad Nacional de Colombia, Bogotá, Colombia \email{\{stoledoc, diusecher, fagonzalezo\}@unal.edu.co}}
\maketitle              
\begin{abstract}
Prostate cancer (PCa) is one of the most common and aggressive cancers worldwide. The Gleason score (GS) system is the standard way of classifying prostate cancer and the most reliable method to determine the severity and treatment to follow. The pathologist looks at the arrangement of cancer cells in the prostate and assigns a score on a scale that ranges from 6 to 10. Automatic analysis of prostate whole-slide images (WSIs) is usually addressed as a binary classification problem, which misses the finer distinction between stages given by the GS. This paper presents a probabilistic deep learning ordinal classification method that can estimate the GS from a prostate WSI. Approaching the problem as an ordinal regression task using a differentiable probabilistic model not only improves the interpretability of the results, but also improves the accuracy of the model when compared to conventional deep classification and regression architectures. 
\end{abstract}

\section{Introduction}

Prostate cancer (PCa) is currently the second most common cancer among men in America. Early detection allows for greater treatment options and a greater chance of treatment success, but while there are several methods of initial screening, a concrete diagnosis of PCa can only be made with a prostate biopsy \cite{Faraj2016}. Tissue samples are currently recorded in high-resolution images, called whole-slide images (WSIs). In these images the pathologists analyze the alterations in the stroma and glandular units and, using the Gleason score (GS) system, classify prostate cancer into five progressive levels from 6 to 10 \cite{Li2020}. The higher the grade, the more advanced the cancer. The analysis is mostly a manual task and requires specialized urological pathologists. This specialized staff is not always available, especially in developing countries, and the process is subject to great inter-observer variability \cite{Strom2020}. Therefore, several efforts have been made to develop computer assisted diagnosis systems which may facilitate the work of specialists \cite{Bulten2020}. 

Deep convolutional neural networks (CNN) represent the state of the art in the analysis of visual information, and their implementation in automatic classification models for medical images has been widely studied. However, there is still much research to be done in relation to the diagnostic process in histopathology \cite{Strom2020}. One of the main problems facing the application of deep learning into medical problems is the limited availability of large databases, given the standard required for the successful training of deep learning models. For histopathology, the previous performed studies have been limited to very small data sets or subsets of Gleason patterns \cite{Strom2020}. In addition, deep learning models approach the prostate tissue grading task as a multi-class or even a binary classification of low risk (6-7 GS) vs high risk (8-10 GS) cases \cite{Lara2020}. This has two drawbacks: first, the ordinal information of the grades is not taken into account. Second, the model predictions, usually subject to a softmax activation function, cannot be interpreted as a probability distribution \cite{vaicenavicius19a}, and therefore do not give information about the uncertainty of the predictions which, in safety-critical applications, provides the method with a first level of interpretability.


In this paper we approach the prostate tissue grading as an ordinal regression task. We present the Deep Quantum Measurement Ordinal Regression (DQMOR), a deep probabilistic model that combines a CNN with a differentiable probabilistic regression model, the Quantum Measurement Regression (QMR) \cite{Gonzalez2021}. This approach allows us to:

\begin{enumerate}
    \item Predict posterior probability distributions over the grades range. Unlike other probabilistic methods as Gaussian processes, these are explicit discrete distributions. 
    \item Integrate patch-level posterior distributions into a single whole-slide image distribution in a simple, yet powerful probability-based manner. 
    \item Quantify the uncertainty of the predictions. This enrich the model as a diagnostic support tool, by providing it with a first level of interaction and interpretability of the results.
\end{enumerate}

In order to validate our approach, we compare our performance with state of the art deep learning-based methods \cite{Lara2020}, and with close related classification and regression methods as the Density Matrix Kernel Density Classification (DMKDC) \cite{Gonzalez2021} and Gaussian processes \cite{Cutajar2017} \cite{Rasmussen2006}.

The paper is organized as follows: Section 2 presents a brief overview of the related work. Section 3 presents the theoretical framework of the DQMOR, and Section 4 presents the experimental set up and results. Finally in Section 5 we present the conclusions of this work. 


\section{Related Work}

Classification of prostate cancer images by GS is considered a difficult task even among pathologist, who do not usually agree on their judgment. In recent years, there has been a great research effort to automatically classify PCa. However, most of the previous works focus on classifying prostate WSIs between low and high GS, ignoring the inherent ordinal characteristics of the grading system. 

To train a CNN with WSIs, it is required to divide each image into multiple patches, and then, to summarize the information of the patches by different methods, hence, obtaining a prediction of the WSI. In \cite{JimenezdelToro2017}, the authors classify patches between low, and high GS, utilizing various CNN architectures and summarizing the patches to a WSI by a GS majority vote. Another approach by Tolkach et al. \cite{Tolkach2020} uses a NASNetLarge CNN, and summarizes the GS of the patches by counting the probabilities per class. In Karimi et al. \cite{Karimi2020} they proposed training three CNNs for patches of different sizes, and summarizing the probabilities by a logistic regression. In \cite{Esteban2019}, the authors use Gaussian processes based on granulometry descriptors extracted with a CNN for the binary classification task. Some other CNN architectures for GS grading include a combination of an atrous spatial pyramid pooling and a regular CNN as in \cite{Li2020}, an Inception-v3 CNN with a support vector machine (SVM) as in \cite{Lucas2019}, and a DeepLabV3+ with a MobileNet as the backbone \cite{Khani2019}. 

\section{Deep Quantum Measurement Ordinal Regression}

\begin{figure}
\begin{center}
\includegraphics[width=\textwidth]{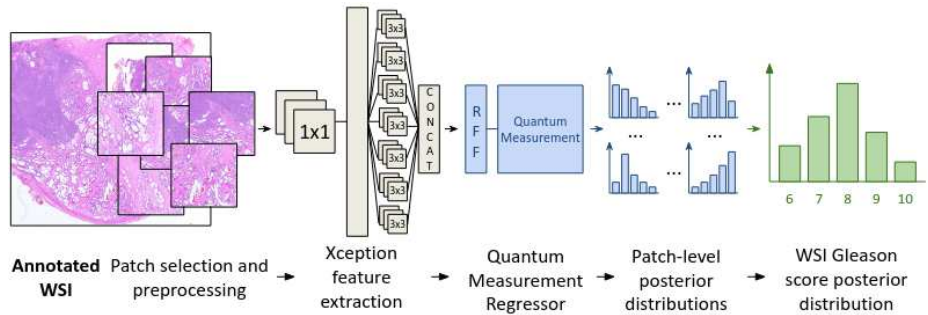}
\caption{Overview of the proposed DQMOR method for prostate tissue grading. A Xception network is used as feature extractor for the image patches. Those features are the input for the QMR regressor model, which yields a posterior probability distribution by patch over the Gleason score grades. Then, those distributions are summarized into a single discrete probability distribution for the WSI.} \label{model}
\end{center}
\end{figure}

The overall architecture of the proposed Deep Quantum Measurement Ordinal Regression (DQMOR) is described in Figure \ref{model}. We use a Xception CNN \cite{chollet2017} as a patch-level feature extractor. The extracted features are then used as inputs for the QMR method \cite{Gonzalez2021}. QMR requires an additional feature mapping from the inputs to get a quantum state-like representation. This is made by means of a random Fourier features approach \cite{Rahimi2009RandomMachines}. The regressor yields a discrete posterior probability distribution at patch level. Then, to predict the GS of a single WSI, we summarize the results of the patches into a single posterior distribution from which we get the final grade and an uncertainty measure.

\subsection{Feature extraction}

We choose as feature extractor the model presented in \cite{Lara2020}, which is publicly available, and consists of an Xception network trained on ImageNet and fine-tuned on prostate tissue image patches. This network was originally used for an automatic information fusion model for the automatic binary (low-high) classification of WSIs. Taking the output of the last average pooling layer of the model we got a 2048-dimensional vector representing each image patch. 

\subsection{Quantum Measurement Regression}

QMR addresses the question on how to use density matrices for regression problems, using random features to encode the inputs in a quantum state-like representation. The model works as a non-parametric density estimator \cite{Gonzalez2021}.

\subsubsection{Random Fourier Features (RFF)}



The RFF method \cite{Rahimi2009RandomMachines} creates a feature map of the data $\mathbf{z(x)}: \mathbb{R}^n \rightarrow \mathbb{R^D}$ in which the dot product of the samples in the $\mathbb{R}^D$ space approximates a shift invariant kernel $k(\mathbf{x} - \mathbf{y})$. The method works by sampling i.i.d. $w_1, \cdots, w_D \in \mathbb{R}^n$ from a probability distribution $p(w)$ given by the Fourier transform of $k(\mathbf{x} - \mathbf{y})$, and sampling i.i.d. $b_1, \cdots, b_D \in \mathbb{R}$ from an uniform distribution in $[0, 2\pi]$. In our context, the shift invariant kernel is the Radial Basis Function (RBF) given by, $\text{k}_\text{RBF}(\mathbf{x}-\mathbf{y})=e^{-\gamma\|\mathbf{x}-\mathbf{y}\|^2}$, where gamma $\gamma$ and the number $D$ of RFF components are hyper-parameters of the models. In our model the RFF works as an embedding layer that maps the features from the Xception module to a representation space that is suitable for the quantum measurement regresion layer. 

\subsubsection{Quantum Measurement Regression (QMR)}

QMR  \cite{Gonzalez2021} is a differentiable probabilistic regression model that uses a density matrix, $\rho_{\text{train}}$, to represent the joint probability distribution of inputs and labels. A QMR layer receives a RFF encoded input sample $\ket{\psi_x}$, and then builds a prediction operator $\pi = \ket{\psi_x}\bra{\psi_x}\otimes \text{Id}_{\mathcal{H}_\mathcal{Y}}$ where $\text{Id}_{\mathcal{H}_\mathcal{Y}}$ is the identity operator in $\mathcal{H}_\mathcal{Y}$, the representation space of the labels. Inference is made by performing a measurement on the training density matrix $\rho_{\text{train}}$:

\begin{equation}
    \rho = \frac{\pi\rho_{\text{train}}\pi}{\text{Tr}[\pi\rho_{\text{train}}\pi]}.
\end{equation}

Then a partial trace $\rho_\mathcal{Y} = \text{Tr}_\mathcal{X}[\rho]$ is calculated, which encodes in $\rho_{\mathcal{Y}rr}$, with $r \in \{0, \dots, N-1\}$, the posterior probability  over the labels. The expected value represents the final prediction $\hat{y}=\sum_{r=0}^{N-1}r\rho_{\mathcal{Y}rr}$.

A gradient-based optimization is allowed by a spectral decomposition of the density matrix, $\rho_\text{train} = V^\dagger\Lambda V$, in which the number of eigen-components of the factorization is a hyper-parameter of the model. The model is trained by minimizing a  mean-squared-error loss function with a variance term whose relative importance is controlled by hyper-parameter $\alpha$:

\begin{equation}\label{eq:loss}
    L = \sum (y-\hat{y})^2 + \alpha \sum_{r} \rho_{\mathcal{Y}rr} (\hat{y}-r)^2.
\end{equation}

\subsection{WSIs predictions}

Since the training of the model is performed at patch level, the evaluation can be done at such level and at the level of the WSI. To get a prediction for a whole image, we explored two approaches: a majority vote procedure (MV), and a probability vote procedure (PV). In the majority vote, the prediction for an image is decided according to the grade with the highest number of predictions among the patches of the image. And in the probability vote, since each patch can be associated with a probability distribution, the normalized summation yields a distribution for the whole image. More formally, thanks to the law of total probability,  given an image $I$, composed by $n$  patches, each patch denoted by $p_i$, the posterior probability of the grade $r$ is, 

\begin{equation}
    P(r|I) =
    \dfrac{P(r,I)}{P(I)} =
    \dfrac{\sum_{i=1}^nP(r|p_i,I)P(p_i|I)P(I)}{P(I)} =
    \frac{1}{n}\sum_{i=1}^nP(r|p_i).\label{eq:2}
\end{equation}

The final prediction value thus corresponds to the grade with highest probability. In the DQMOR method, one can predicts the expected value of the distribution, but instead, the predictions at patch level were deduced from the probability of each grade per patch $P(r|p_i)$, and at WSI level by MV and PV.





\section{Experimental Evaluation}

\subsection{Dataset}

We use images from the TCGA-PRAD data set, which contains samples of prostate tissue with GS from 6 to 10. This data set is publicly available via The Cancer Genome Atlas (TCGA) \cite{JimenezdelToro2017}. In order to directly compare our results with our baseline \cite{Lara2020} we use the same subset and partition consisting of 141 cases for training, 48 for validation and 46 for testing.

\subsection{Experimental Set Up}

The feature extraction model is publicly available and the augmentation procedure and training details are described in \cite{Lara2020}. For the QMR, hyper-parameter tuning of the model was performed by generating 25 different random configurations. As result, we created an embedding with 1024 RFF components, 32 eigenvalues and $\gamma$ was set to $2^{-13}$. For the loss function (See eq. (\ref{eq:loss})), $\alpha$ was set at $0.4$, and we set a learning rate of $6\times10^{-5}$. 




Two extensions of the feature extractor model were set up as baseline for this work. The first dense layer classifier (DLC-1) consisted on 1024 neurons with ReLU as the activation function and a dropout of $0.2$, followed by 5 neurons with a soft-max activation function for the output, and the learning rate was set to $10^{-7}$, as in the baseline \cite{Lara2020}. The second classifier (DLC-2) had two dense layers of $100$ and $50$ neurons with ReLU activation functions and dropouts of $0.2$, connected to 5 neurons with a softmax activation function, and the learning rate was set to $10^{-3}$. 

We also explored two closely related methods to QMR: Density Matrix Kernel Density Classification (DMKDC)  \cite{Gonzalez2021} and Gaussian processes. DMKDC is a differentiable classification method, which applies a RFF feature map to the input sample, and then computes the expected value of the input with a density matrix of each class, returning a posterior probability distribution, which can be optimized with a categorical cross entropy loss function. As with QMR, a hyper-parameter random search was performed. We created an embedding with 1024 RFF components, and 32 eigenvalues. $\gamma$ was set up at $2^{-13}$, and we set a learning rate of $5\times10^{-3}$. All the previous experiments were performed in Python using the publicly available Keras-based implementation presented in \cite{Gonzalez2021}.

On the other hand, Gaussian processes (GP) \cite{Rasmussen2006} are another powerful Bayesian approach to regression problems. By means of a kernel covariance matrix, the GP calculates and updates iteratively the probability distribution of all the functions that fit the data, optimizing in the process the kernel parameters. In our case we set the kernel as the RBF. The prediction process consist in marginalizing the learned Gaussian distribution, whose mean would be the actual prediction value, and its standard deviation an uncertainty indicator. We performed experiments with GP using the Scikit-Learn implementation in Python. We also explored deep Gaussian processes (DGP), using the implementation proposed in \cite{Cutajar2017}, which also uses RFF to approximate the covariance function. For those experiments, another hyper-parameter random search was made, finally setting the number of RFF at 1024 and the learning rate at $2\times10^{-12}$ in a single layer schema.

\subsection{Results and Discussion}


\subsubsection{Ordinal Regression}

To measure the performance of an ordinal regression method implies to take into account the severity of the misclassified samples. Therefore, in addition to  accuracy (ACC) and macro f1 score, we also measured mean absolute error (MAE) on the test partition, at patch level and WSI level. WSI scores were summarized by a MV and PV. The prediction methods at WSI-level were also applied to the baseline models. In the dense layers classifiers from the softmax output,  as in \cite{Tolkach2020}. In the DMKDC, the prediction methods were easily applied because the model outputs a probability distribution. For GP and DGP only MV was calculated, since we have no access to an explicit discrete posterior distribution. The results are reported in Table \ref{tab_results_patch} and Table \ref{tab_results_wsi}.

\begin{table}
\begin{center}
\caption{Patch-level results of the two dense layers classifiers models DCL-1, DCL-2, Gaussian processes GP, DGP, and density matrix-based models DMKDC, DQMOR. Mean and standard deviation of accuracy, macro f1 score and mean absolute error (MAE) are reported over 10 trials. }\label{tab_results_patch}
    \begin{tabular}{|c|c|c|c|}
        \hline
        Method & Accuracy & Macro F1 & MAE\\
        \hline
        DLC-1 \cite{Lara2020}& $0.530 \pm 0.001$ & \textbf{0.314}$\pm0.001$ & $0.786\pm0.002$\\
        DLC-2 \cite{Lara2020}& $0.542 \pm 0.005$ & $0.296 \pm 0.007$ & $0.780 \pm 0.009$ \\
        GP \cite{Rasmussen2006}& $ 0.399 \pm 0.000$ & $0.255 \pm 0.000$ & $0.777 \pm 0.000$ \\
        DGP \cite{Cutajar2017}& $ 0.265 \pm 0.001$ & $0.169 \pm 0.000$ & $1.013 \pm 0.000$ \\
        DMKDC \cite{Gonzalez2021} & \textbf{0.546} $\pm 0.002$ & $0.305 \pm 0.006$& $0.775 \pm 0.007$ \\
        \hline
        \textbf{DQMOR}  & $0.477 \pm 0.006$ & $0.293 \pm 0.003$  &  \textbf{0.732}$\pm 0.005 $ \\
        
        \hline
    \end{tabular}
\end{center}
\end{table}

In terms of accuracy at patch level, the DMKDC model obtained the highest results. The best accuracy at WSI level was reached with the DQMOR model with probability vote. The DQMOR also obtained the least mean absolute errors at patch and WSI levels, showing that the model take advantage of the probability distributions and the inherent ordinality of the GS grades.

\begin{table}
\begin{center}
\caption{WSI-level results. For each model, two summarization procedures are applied, majority vote (MV) and probability vote (PV). Mean and standard deviation of accuracy, macro f1 score and mean absolute error (MAE) are reported over 10 trials.  }\label{tab_results_wsi}
    \begin{tabular}{|c|c|c|c|}
        \hline
        Method & Accuracy & Macro F1 & MAE\\
        \hline
        DLC-1 MV \cite{Lara2020} & $0.543\pm0.000$ & $0.292\pm0.000$ & $0.826\pm0.000$\\
        DLC-2 MV \cite{Lara2020} & \textbf{0.548}$\pm0.009$ & $0.300\pm0.016$ & $0.822\pm0.009$ \\
        GP MV \cite{Rasmussen2006}  & $0.391\pm0.000$ & $0.233\pm0.000$ & $0.739\pm0.000$ \\
        DGP MV \cite{Cutajar2017}  & $0.174\pm0.000$ & $0.059\pm0.000$ & $0.935\pm0.000$ \\
        DMKDC MV \cite{Gonzalez2021} & $0.546\pm0.002$ & $0.296\pm0.012$ & $0.824\pm0.006$ \\
        \hline
        \textbf{DQMOR MV} & $0.513\pm0.014$ & \textbf{0.306}$\pm0.010$ & \textbf{0.713}$\pm0.027$ \\
        \hline
        DLC-1 PV \cite{Lara2020}& $0.543\pm0.000$ & $0.292\pm0.000$ & $0.826\pm0.000$ \\
        DLC-2 PV \cite{Lara2020}& $0.550\pm0.005$ & $0.304\pm0.018$ & $0.820\pm0.010$ \\
        DMKDC PV \cite{Gonzalez2021}& $0.546\pm0.002$ & $0.296\pm0.012$ & $0.824\pm0.006$ \\
        \hline
        \textbf{DQMOR PV} & \textbf{0.567}$\pm0.021$ & \textbf{0.345}$\pm0.014$ & \textbf{0.730}$\pm0.024$ \\
        
        \hline
    \end{tabular}
\end{center}
\end{table}

\subsubsection{Uncertainty Quantification}

\begin{figure}
\begin{center}
\includegraphics[width=0.5\textwidth]{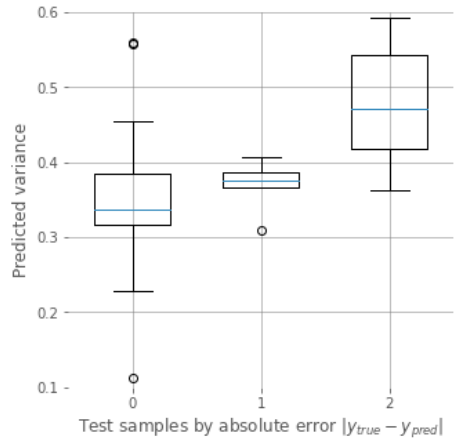}
\caption{Boxplot of the predicted variance on test samples at WSI-level, grouped by absolute classification error $|y_{true}-y_{pred}|$.} \label{fig1}
\end{center}
\end{figure}

Beyond the classification and regression performance of the methods, DQMOR allows an uncertainty quantification based on the variance of the predicted distribution. We analyzed the statistical behaviour of the predicted variance on the test set at WSI-level, grouping the samples according to the absolute error $|y_{true}-y_{pred}|$. As expected, DQMOR predicts low uncertainty levels on well classified samples when compared with the miss-classified samples (see Figure \ref{fig1}). In fact, the greater the absolute error, the greater the uncertainty. This attribute provides the method with an interpretable mean for the specialist, who may decide whether to trust or not in the model prediction.

\section{Conclusions}
In this work we approached the prostate tissue grading as an ordinal regression task. We combined the representational power of deep learning with the Quantum Measurement Regression method, which uses density matrices and random features to build a non-parametric density estimator.


The results on classification and regression metrics show that at WSI-level, DQMOR outperforms similar probabilistic classification and regression methods, as well as extension of the deep base model used for feature extraction. Regarding the analysis of the predicted uncertainty, we showed that DQMOR allows the identification of misclassified examples, and that the higher the misclassification error, the higher the uncertainty. This is a highly valued ability in medical applications, where the aim is to prevent false positives and especially false negatives in a diagnostic processes.

Overall we demonstrate that unlike single deep learning architectures and standard classification models, the combination of CNN's and QMR allows us to use the ordinal information of the disease grades, and provides a better theoretical framework to combine the patch-level inference into a single WSI prediction. 




%
%
%
\bibliographystyle{splncs04}
%

\bibliography{bib}

\begin{thebibliography}{10}
\providecommand{\url}[1]{\texttt{#1}}
\providecommand{\urlprefix}{URL }
\providecommand{\doi}[1]{https://doi.org/#1}

\bibitem{Bulten2020}
Bulten, W., Pinckaers, H., van Boven, H., Vink, R., de~Bel, T., van Ginneken,
  B., van~der Laak, J., {Hulsbergen-van de Kaa}, C., Litjens, G.: {Automated
  deep-learning system for Gleason grading of prostate cancer using biopsies: a
  diagnostic study}. The Lancet Oncology  \textbf{21}(2),  233--241 (2020).
  \doi{10.1016/S1470-2045(19)30739-9},
  \url{http://dx.doi.org/10.1016/S1470-2045(19)30739-9}

\bibitem{chollet2017}
Chollet, F.: Xception: Deep learning with depthwise separable convolutions
  (2017)

\bibitem{Cutajar2017}
Cutajar, K., Bonilla, E.V., Michiardi, P., Filippone, M.: {Random feature
  expansions for Deep Gaussian Processes}. 34th International Conference on
  Machine Learning, ICML 2017  \textbf{2},  1467--1482 (2017)

\bibitem{Esteban2019}
Esteban, {\'{A}}.E., L{\'{o}}pez-P{\'{e}}rez, M., Colomer, A., Sales, M.A.,
  Molina, R., Naranjo, V.: {A new optical density granulometry-based descriptor
  for the classification of prostate histological images using shallow and deep
  Gaussian processes}. Computer Methods and Programs in Biomedicine
  \textbf{178},  303--317 (2019). \doi{10.1016/j.cmpb.2019.07.003}

\bibitem{Faraj2016}
Faraj, S.F., Bezerra, S.M., Yousefi, K., Fedor, H., Glavaris, S., Han, M.,
  Partin, A.W., Humphreys, E., Tosoian, J., Johnson, M.H., Davicioni, E.,
  Trock, B.J., Schaeffer, E.M., Ross, A.E., Netto, G.J.: {Clinical validation
  of the 2005 isup gleason grading system in a cohort of intermediate and high
  risk men undergoing radical prostatectomy}. PLoS ONE  \textbf{11}(1),  1--13
  (2016). \doi{10.1371/journal.pone.0146189}

\bibitem{Gonzalez2021}
Gonz{\'{a}}lez, F.A., Gallego, A., Toledo-Cort{\'{e}}s, S.,
  Vargas-Calder{\'{o}}n, V.: {Learning with Density Matrices and Random
  Features}  (2021), \url{http://arxiv.org/abs/2102.04394}

\bibitem{JimenezdelToro2017}
{Jim{\'{e}}nez del Toro}, O., Atzori, M., Ot{\'{a}}lora, S., Andersson, M.,
  Eur{\'{e}}n, K., Hedlund, M., R{\"{o}}nnquist, P., M{\"{u}}ller, H.:
  {Convolutional neural networks for an automatic classification of prostate
  tissue slides with high-grade Gleason score}. Medical Imaging 2017: Digital
  Pathology  \textbf{10140},  101400O (2017). \doi{10.1117/12.2255710}

\bibitem{Karimi2020}
Karimi, D., Nir, G., Fazli, L., Black, P.C., Goldenberg, L., Salcudean, S.E.:
  {Deep Learning-Based Gleason Grading of Prostate Cancer from Histopathology
  Images - Role of Multiscale Decision Aggregation and Data Augmentation}. IEEE
  Journal of Biomedical and Health Informatics  \textbf{24}(5),  1413--1426
  (may 2020). \doi{10.1109/JBHI.2019.2944643}

\bibitem{Khani2019}
Khani, A.A., {Fatemi Jahromi}, S.A., Shahreza, H.O., Behroozi, H., Baghshah,
  M.S.: {Towards Automatic Prostate Gleason Grading Via Deep Convolutional
  Neural Networks}. 5th Iranian Conference on Signal Processing and Intelligent
  Systems, ICSPIS 2019 (December),  18--19 (2019).
  \doi{10.1109/ICSPIS48872.2019.9066019}

\bibitem{Lara2020}
Lara, J.S., Contreras~O., V.H., Ot{\'a}lora, S., M{\"u}ller, H., Gonz{\'a}lez,
  F.A.: Multimodal latent semantic alignment for automated prostate tissue
  classification and retrieval. In: Martel, A.L., Abolmaesumi, P., Stoyanov,
  D., Mateus, D., Zuluaga, M.A., Zhou, S.K., Racoceanu, D., Joskowicz, L.
  (eds.) Medical Image Computing and Computer Assisted Intervention -- MICCAI
  2020. pp. 572--581. Springer International Publishing, Cham (2020)

\bibitem{Li2020}
Li, Y., Huang, M., Zhang, Y., Chen, J., Xu, H., Wang, G., Feng, W.: {Automated
  Gleason Grading and Gleason Pattern Region Segmentation Based on Deep
  Learning for Pathological Images of Prostate Cancer}. IEEE Access
  \textbf{8},  117714--117725 (2020). \doi{10.1109/ACCESS.2020.3005180}

\bibitem{Lucas2019}
Lucas, M., Jansen, I., Savci-Heijink, C.D., Meijer, S.L., de~Boer, O.J., van
  Leeuwen, T.G., de~Bruin, D.M., Marquering, H.A.: {Deep learning for automatic
  Gleason pattern classification for grade group determination of prostate
  biopsies}. Virchows Archiv  \textbf{475}(1),  77--83 (2019).
  \doi{10.1007/s00428-019-02577-x}

\bibitem{Rahimi2009RandomMachines}
Rahimi, A., Recht, B.: {Random features for large-scale kernel machines}. In:
  Advances in Neural Information Processing Systems 20 - Proceedings of the
  2007 Conference (2009)

\bibitem{Rasmussen2006}
Rasmussen, C.E., Williams, C.K.I.: {Gaussian processes for machine learning.}
  The MIT Press (2006). \doi{10.1142/S0129065704001899},
  \url{http://www.gaussianprocess.org/gpml/chapters/RW.pdf}

\bibitem{Strom2020}
Str{\"{o}}m, P., Kartasalo, K., Olsson, H., Solorzano, L., Delahunt, B.,
  Berney, D.M., Bostwick, D.G., Evans, A.J., Grignon, D.J., Humphrey, P.A.,
  Iczkowski, K.A., Kench, J.G., Kristiansen, G., van~der Kwast, T.H., Leite,
  K.R., McKenney, J.K., Oxley, J., Pan, C.C., Samaratunga, H., Srigley, J.R.,
  Takahashi, H., Tsuzuki, T., Varma, M., Zhou, M., Lindberg, J., Lindskog, C.,
  Ruusuvuori, P., W{\"{a}}hlby, C., Gr{\"{o}}nberg, H., Rantalainen, M.,
  Egevad, L., Eklund, M.: {Artificial intelligence for diagnosis and grading of
  prostate cancer in biopsies: a population-based, diagnostic study}. The
  Lancet Oncology  \textbf{21}(2),  222--232 (2020).
  \doi{10.1016/S1470-2045(19)30738-7}

\bibitem{Tolkach2020}
Tolkach, Y., Dohmg{\"{o}}rgen, T., Toma, M., Kristiansen, G.: {High-accuracy
  prostate cancer pathology using deep learning}. Nature Machine Intelligence
  \textbf{2}(7),  411--418 (jul 2020). \doi{10.1038/s42256-020-0200-7},
  \url{https://www.nature.com/articles/s42256-020-0200-7}

\bibitem{vaicenavicius19a}
Vaicenavicius, J., Widmann, D., Andersson, C., Lindsten, F., Roll, J.,
  Sch\"{o}n, T.: Evaluating model calibration in classification. In: Chaudhuri,
  K., Sugiyama, M. (eds.) Proceedings of Machine Learning Research. Proceedings
  of Machine Learning Research, vol.~89, pp. 3459--3467. PMLR (16--18 Apr
  2019), \url{http://proceedings.mlr.press/v89/vaicenavicius19a.html}

\end{thebibliography}

\end{document}